\documentclass{article}

\usepackage{arxiv}

\usepackage[utf8]{inputenc} 
\usepackage[T1]{fontenc}    
\usepackage{hyperref}       
\usepackage{url}            
\usepackage{booktabs}       
\usepackage{amsfonts}       
\usepackage{nicefrac}       
\usepackage{multicol}       
\usepackage[ruled, vlined]{algorithm2e}

\usepackage{microtype}      
\usepackage{lipsum}
\usepackage{graphicx}
\graphicspath{ {./images/} }
\usepackage{float}

\title{SDL: New data generation tools for full-level annotated document layout}

\author{
 Nguyen Truong Son \\
  School of Environment and Society\\
  Tokyo Institute of Technology\\
  Tokyo, Japan \\
  \texttt{nguyen.s.ab@m.titech.ac.jp} \\
}

\begin{document}
\maketitle
\begin{abstract}
We present a novel data generation tool for document processing. The tool focuses on providing maximal level of visual information in a normal type document, ranging from character position to paragraph-level position. It also enables working with a large dataset on low-resource languages as well as providing a mean of processing thorough full-level information of documented text. The data generation tools come with a dataset of 320000 Vietnamese synthetic document images and an instruction to generate a dataset of similar size on other  languages. The repository can be found at: \url{https://github.com/tson1997/SDL-Document-Image-Generation}
\end{abstract}

\begin{multicols*}{2}[]

\section{Introduction}
Optical Character Recogition (OCR) is the problem of reading and extracting information from an image. In order to extract correct information in the image, multiple level detection were made. They range from character-level detection to paragraph level detection. Such set of document structuring problems is called Document Layout Analysis (DLA). The goal of DLA is to decompose an image of document into multiple parts, whose information are then further processed or filtered based on their usefulness. 

With the power of deep learning, DLA can solve the detection problems by using a good object detection model, such as Faster RCNN \cite{fasterrcnn}, or Mask RCNN \cite{he2017}. Hence, the problems left is how to obtain a good dataset to train such object detection model. \cite{Gupta16, long2020unrealtext} proposes ways to generate synthetic scene text image. Such synthetic datasets yield good variety of scene text image with text blending into background. However, we cannot use those datasets directly to train a model for decomposing a document image into level component that are bigger than word-level, since there are differences in distribution between document text and scene text data.

On the other hand, \cite{zhong2019publaynet} provided a way to process documented data from the Internet and provided their resulted PublayNet dataset of more than 300000 images and corresponding label. However, PublayNet dataset has the limitation on label-level (only big component such as paragraph, table, list), and limitation on language domain (only in English). 

Despite the fact that there are many open-source for both scene-text dataset, at the time of this paper, there is no dataset that covers a full level of label from character up to big page component like paragraph. There is a need for such data providing an end-to-end solution, where all components ranging from smallest like character to biggest like paragraph would be covered thoroughly. In addition, there is also a need for data providing flexibility in languages, especially low-resource language where collecting big dataset is not feasible.

In this work, we are going to provide a new method to generate synthetic data, with the main focus on multi-level text component, and possible to generate in multiple languages. In short, our main contributions are:
\begin{itemize}
    \item First algorithm to generate synthetic visual dynamic document layout,
    \item An algorithm that fits in multiple languages document, even low-resource language.
    \item First big document dataset in Vietnamese with biggest number of images and biggest number of component levels, ranging from character-level to paragraph-level.
\end{itemize}
\begin{figure*} 
    \centering
    \includegraphics[width=0.45\textwidth]{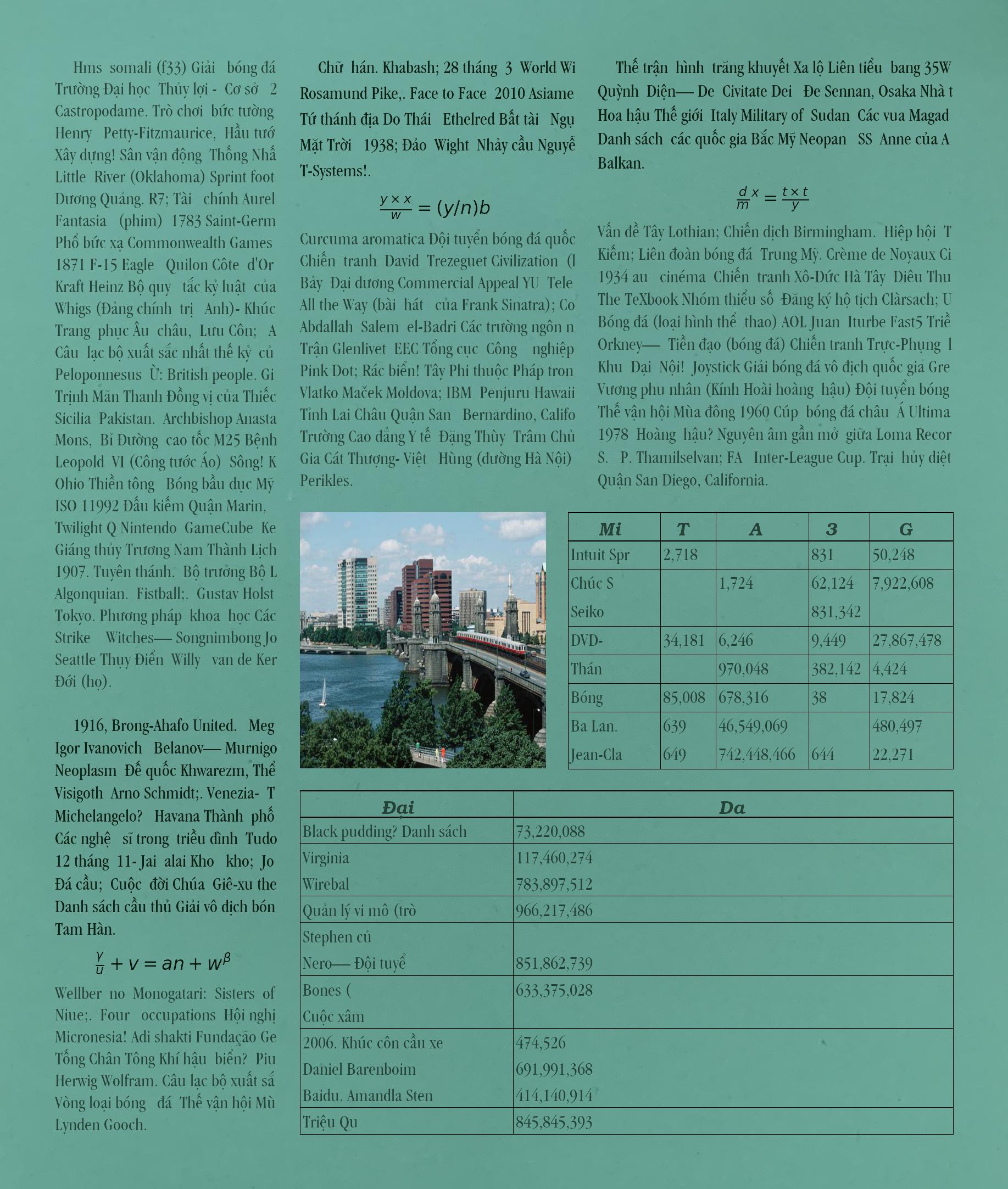}
    \includegraphics[width=0.45\textwidth]{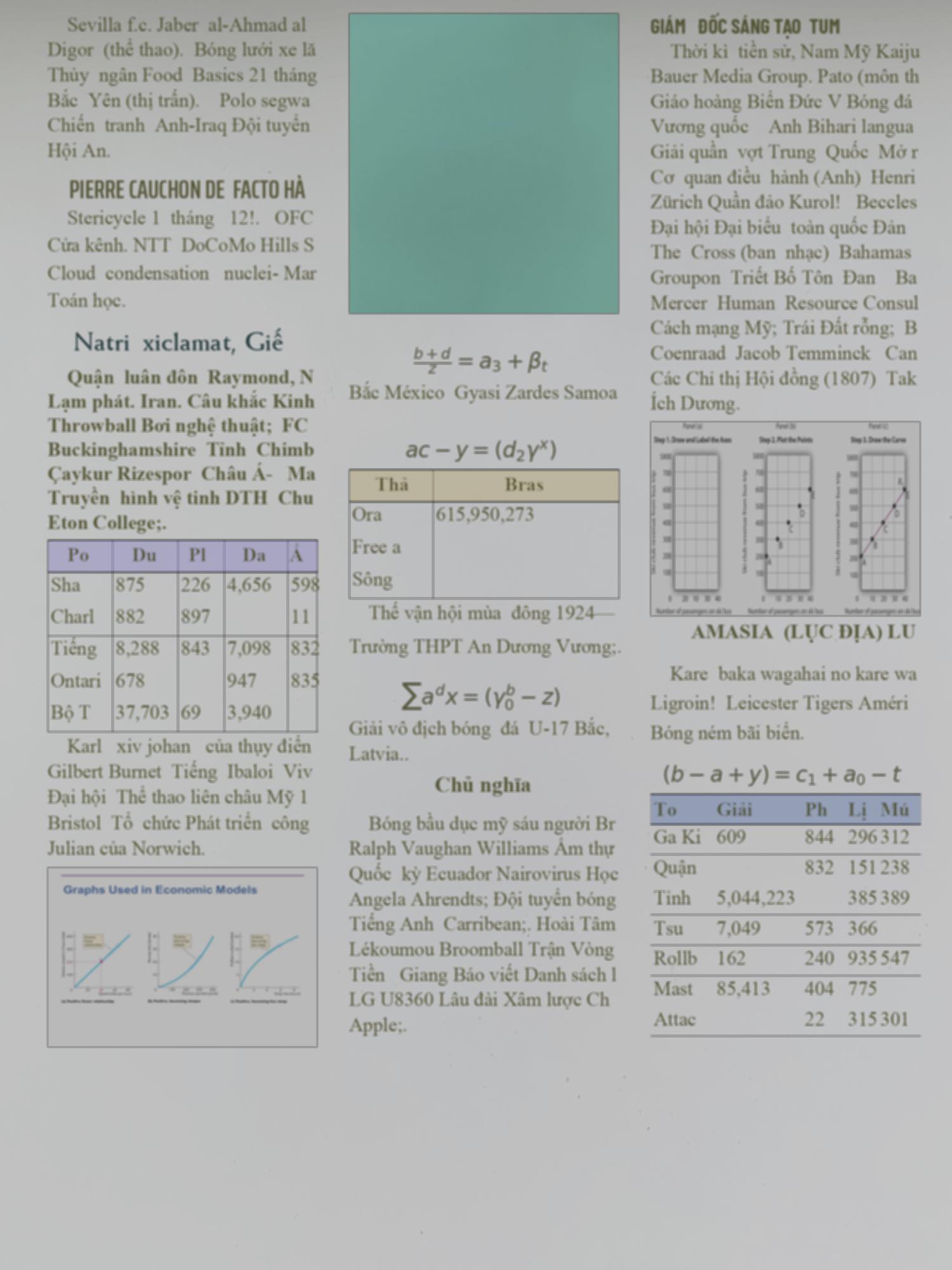}
    \caption{Sample images from the dataset. Left: Flexible layout; Right: Fixed columns layout.}
    \label{fig:fig1}
\end{figure*}
\section{Related Work}
There are two main approaches on gathering big datasets for document layout or any text document dataset in general. The first set of approach is to generate synthetic data that is close to real data that we are going to use. The second set of approach is to crawl real data and either label it manually or automatically.
\subsection{Synthetic Text dataset}
Throughout the years, as deep learning becomes the mainstream in many tasks including optical character recognition, large amount of text images data for training deep model has become more and more important. However, there are many low-resource languages where we cannot crawl much of the data from open source like the Internet. Hence, synthetically generating text dataset comes naturally as an inevitable option for those languages. As far as we know, there are several attempts to generate synthetic data for optical character recognition.

\cite{Gupta16} provided the first synthetic scene text dataset, with most number of images. The datasets help a lot in pretraining a good model for further fine-tuning on many text detection dataset. In addition, it is well noted that many people used \cite{Gupta16} open-source code to generate their own custom dataset in multiple languages. It is noted that up to this day, there is no real dataset that can match the number of data instances that SynthText provided as well as the capacity it provided to generate text in multiple languages. 

Another approach by \cite{long2020unrealtext} combined the power of 3D Unreal Engine in generating persuasive text in the wild data. The data were surreal and can be useful in task like scene-text recognition. It also provides code that people can generate their own custom data and use for many languages.

There are also frameworks that focuses on generating synthetic text document. However, to the best of our knowledge, there has been no visual synthetic dataset made for document prior to ours.

\subsection{Document Layout dataset}

There are several open source dataset on document layout. They range from detecting multiple components of the document at the same level like {\it paragraph, table, list, etc.} to small text level such as word. For example, \cite{inbook} provided AR IIIT13k focusing on table detection, but contain several other document components. [author] provided DocBank, claiming the largest dataset that contains both document visual information and textual information. PublayNet [reference] is another dataset that is widely used for pretraining document layout detection due to its utility and usage of real data. 

There are, however, disadvantages behind such data collecting approaches. First, the data is domain limited. For example, DocBank and Publaynet works well only on dataset that are close to medical document or research document when they do zero-shot detection. Second, crawling data is not possible if the targeted language and dataset are either low-resource or private resources. It means that for data whose visual are far from English like Arabic or Thai, those datasets are not very effective.
\section{Methods}
\label{sec:methods}
\begin{figure*}
    \centering
    \includegraphics[width=0.8\textwidth]{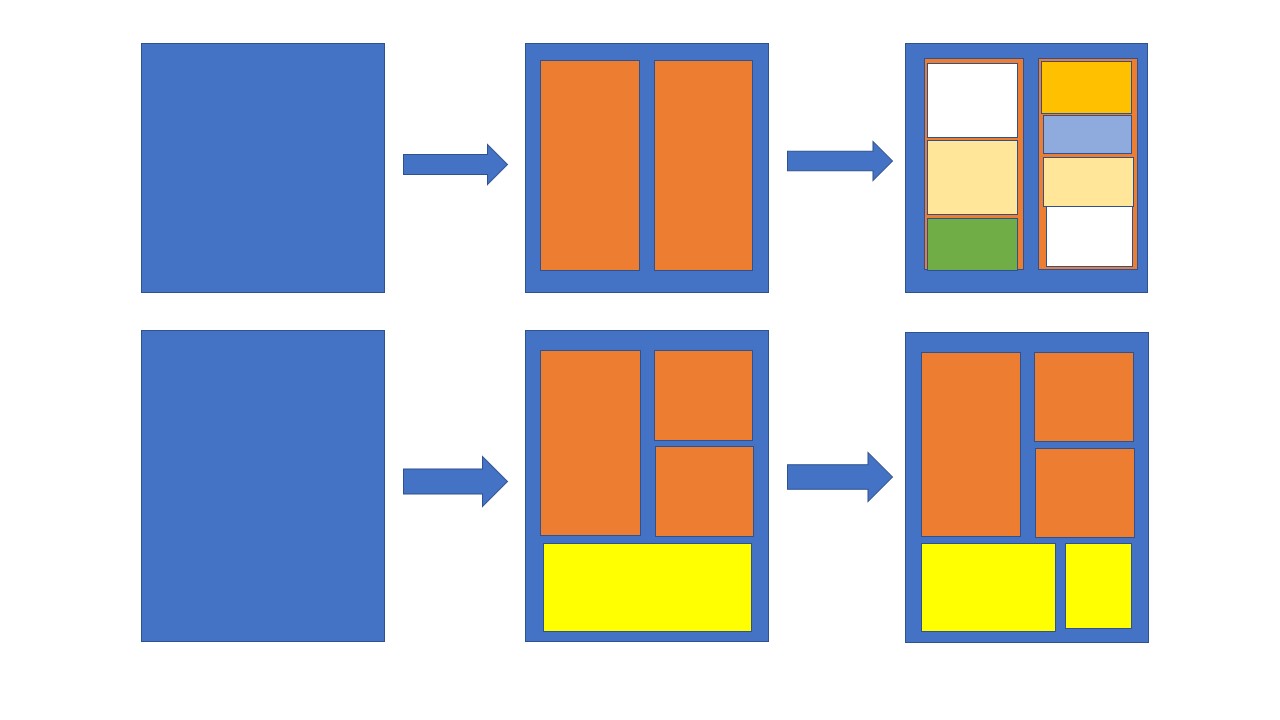}
    \caption{Layout generation algorithm illustration. Up: fixed column; Down: Flexible layout}
    \label{fig:fig2}
\end{figure*}

We aimed at filling the document with 5 forms of document components: "Title", "Paragraph", "Table", "Image", "Formula". Thus, we divided our algorithm into a 2-step pipeline. First, we generate random math formula and crawl image data for "Formula" and "Image" class of component. Second, we use one of the two algorithms discussed later to divide a whole page into multiple regions to fill document components into. Finally, we use Pillow library to fill text and fill image component into a background paper. It would result in a large number of document. 
We provided two options for data generation. The first option is to generate document with layout that are divided into columns. The second option is to make document with more flexibility: we do not have any limitation on either rows or columns. [Figure \ref{fig:fig2}]

\subsection{Fixed columns layout algorithm}
\begin{algorithm}[H]
\SetAlgoLined
\KwResult{Images and Labels}
Initialize numColumns;

 \While{generating}{
  {\it randomize} n in range(1,numColumns);
  
  {\em divide} a page into n columns;
  
  \For{each column}{
    {\it randomize} column break position;
    
    {\it divide} the column by break points;
    
    fill text component into divided regions;
  }
}
 \caption{Fixed layout algorithms}
 \label{fig:fig3}
\end{algorithm}

With this option, we focus on the range of document whose layouts are only divided into specific number of columns. This was inspired by the style that many research document was presented in. By training the detection algorithm on this generated dataset, we can tell the algorithm to separate paragraph in vertical way even though some of the text columns can be close to each other. In addition, such dataset distribution would have a similar distribution to PublayNet. Thus, we can be use it as an alternative version of PublayNet on low-resource languages.

\subsection{Flexible layout generation algorithm}
The second option is to develop an algorithm that can freely divide a page into multiple sections/ regions, and fill many information in such regions. Due to the wide variety of document data structure in the wild, this option helps model to explore a wider set of data, thus can represent document model in a better manner. Experiments on this was done and the differences of pre-training models on flexible layout SDL and pre-training models on other limited domain dataset would be shown in the next section.
\begin{algorithm}[H]
\SetAlgoLined
\KwResult{Images and Labels}
 Initialize minimal area of components;
 
 \While{component area not too small}{
 {\it randomize} direction to divide;
  
  {\it divide} component area into 2 parts along the direction;
}\ 
\For{each divided region}{
  fill text component into the region;
}

 \caption{Flexible layout algorithm}
 \label{fig:fig4}
\end{algorithm}

\paragraph{Implementation details}
For the Vietnamese synthetic document dataset, we use fontsize ranging from 18 to 31, with the height and width of the image ranging from 1500 to 2500. The text color in the dataset are mainly black with a small variance. The text are filled into 5 different categories: Paragraph, Title, Table, Figure, and Formula. For Paragraph, Title and Table component, text annotations are divided into 4 different levels: component-level, line-level, word-level and character-level.

\section{Future work}
There are several ways that we would like to proceed with our data generation tool. First, multi-level text detection models such as TextFuseNet \cite{ijcai2020-72} or CRAFT \cite{baek2019character} can be implemented on our dataset to get a better weight on document model. Second, we can develop a light-weight segmentation model that would take advantages of a multi-level annotations in our dataset. Finally, we would like to have a collaboration on generating dataset in multiple languages other than Vietnamese or English, which would bring benefits to more people who are using low-resource languages.

\bibliographystyle{unsrt}  
\bibliography{references}  

\end{multicols*}
\end{document}